\documentclass[a4paper,twoside]{article}

\usepackage{epsfig}
\usepackage{subcaption}
\usepackage{calc}
\usepackage{amssymb}
\usepackage{amstext}
\usepackage{amsmath}
\usepackage{amsthm}
\usepackage{multicol}
\usepackage{pslatex}
\usepackage{algorithm2e}
\usepackage[bottom]{footmisc}

\usepackage{natbib}
\usepackage{xcolor}
\usepackage{hyperref}

\definecolor{darkblue}{rgb}{0, 0, 0.5}
\hypersetup{colorlinks=true, citecolor=black, linkcolor=darkblue, urlcolor=darkblue}

\renewcommand\cite{\citep}  

\usepackage{graphicx}
\usepackage[nolist]{acronym}
\usepackage{multirow}
\begin{acronym}
    \acro{rq}[RQ]{research question}
    \acroplural{rq}[RQs]{research questions}
    \acro{nlp}[NLP]{natural language processing}
    \acro{llm}[LLM]{large language model}
    \acro{lora}[LoRA]{low-rank adaptation}
    \acroplural{llm}[LLMs]{large language models}
    \acro{plm}[PLM]{pre-trained language model}
    \acroplural{plm}[PLMs]{pre-trained language models}
    \acro{ai}[AI]{artificial intelligence}
    \acro{qa}[QA]{question answering}
    \acro{cqa}[CQA]{Conversational question answering}
    \acro{kbqa}[KBQA]{question answering over knowledge bases}
    \acro{nlg}[NLG]{Natural language generation}
    \acro{kg}[KG]{knowledge graph}
    \acroplural{kg}[KGs]{knowledge graphs}
\end{acronym}
\usepackage{SCITEPRESS}     

\begin{document}

\title{Evaluating Large Language Models in Semantic Parsing for Conversational Question Answering over Knowledge Graphs}


\author{\authorname{Phillip Schneider\sup{1}\orcidAuthor{0000-0001-9492-2927}, Manuel Klettner\sup{1}, Kristiina Jokinen\sup{2}\orcidAuthor{0000-0003-1229-239X}, Elena Simperl\sup{3}\orcidAuthor{0000-0003-1722-947X}, Florian Matthes\sup{1}\orcidAuthor{0000-0002-6667-5452}}
\affiliation{\sup{1}Department of Computer Science, Technical University of Munich, Germany}
\affiliation{\sup{2}AI Research Center, National Institute of Advanced Industrial Science and Technology, Japan}
\affiliation{\sup{3}King’s College London, Department of Informatics, United Kingdom}
\affiliation{\sup{1}\{phillip.schneider, manuel.klettner, matthes\}@tum.de, \sup{2}kristiina.jokinen@aist.go.jp, \sup{3}elena.simperl@kcl.ac.uk
}}

\keywords{conversational question answering, knowledge graphs, large language models, semantic parsing}

\abstract{Conversational question answering systems often rely on semantic parsing to enable interactive information retrieval, which involves the generation of structured database queries from a natural language input. For information-seeking conversations about facts stored within a knowledge graph, dialogue utterances are transformed into graph queries in a process that is called knowledge-based conversational question answering. This paper evaluates the performance of large language models that have not been explicitly pre-trained on this task. Through a series of experiments on an extensive benchmark dataset, we compare models of varying sizes with different prompting techniques and identify common issue types in the generated output. Our results demonstrate that large language models are capable of generating graph queries from dialogues, with significant improvements achievable through few-shot prompting and fine-tuning techniques, especially for smaller models that exhibit lower zero-shot performance.}

\onecolumn \maketitle \normalsize \setcounter{footnote}{0} \vfill

\section{\uppercase{Introduction}}
\label{sec:introduction}
Conversational search has emerged as a growing area of interest within the information retrieval research field in recent years. This burgeoning search paradigm casts the information-seeking process into multi-turn dialogues with conversational agents. The latter are designed to facilitate exploring and gradually narrowing down the search scope to relevant information items \cite{aliannejadi2021analysing, schneider2023investigating}. A fundamental aspect of these agents is their ability to access data stored in knowledge bases with an inherent semantic structure, such as relational databases or knowledge graphs. Hence, a key challenge lies in bridging the gap between natural language utterances, where users express their information needs, and corresponding formal representations, such as logical forms or executable queries. In the field of \ac{nlp}, the task of semantic parsing focuses on this problem by deriving machine-readable meaning representations given linguistic inputs. \ac{cqa} over knowledge graphs is a specialized facet of conversational search that revolves around responding to user queries in a dialogue given an underlying knowledge graph. Through semantic parsing mechanisms, dialogue utterances are transformed into executable queries to retrieve answer triples from a graph. Semantic parsing regarding question answering has been extensively studied over numerous years, with approaches ranging from rule-based to supervised neural network-based techniques.

With the advent of pre-trained \acp{llm}, the field of \ac{nlp} has witnessed a shift in methodologies. Unlike conventional supervised learning approaches that rely on annotated datasets, \acp{llm} are trained in a self-supervised manner, predicting tokens within vast amounts of unlabeled data. Combined with scaling up model size and training corpora, this approach has demonstrated remarkable emergent capabilities of \acp{llm} and their prowess in multi-task learning \cite{radford2019language}. Given a carefully defined input prompt, \acp{llm} have the advantage of prompt-based learning, or in-context learning, which allows them to perform a range of generative tasks like, for instance, question answering, machine translation, or semantic parsing \cite{liu2023survey}. Owing to their contextual language understanding and versatile language generation capabilities, \acp{llm} have experienced rapid adoption in applications related to conversational search, such as personal voice assistants, web browsers, or enterprise chatbots. 
There has been a growing interest in optimizing \acp{llm} for conversational interactions using instruction fine-tuning and reinforcement learning from human feedback \cite{openai2022chat}. Although \acp{llm} offer tremendous potential, it is crucial to acknowledge their inherent limitations, such as the risk of hallucinating or omitting information and a lack of accountability in high-risk scenarios, with limited transparency of the information sources from which they derive their outputs \cite{ji2023survey}. Therefore, it becomes imperative to ground their generated outputs in verifiable facts contained in knowledge bases, which can be facilitated through semantic parsing.

The goal of our study lies in investigating how \acp{llm} perform in semantic parsing of dialogues for \ac{cqa} over knowledge graphs. To answer the stated question, we systematically compare generated outputs from \acp{llm} of varying sizes and training objectives, with a primary focus on models optimized for conversational interaction. Based on an extensive benchmark dataset, we evaluate the models' performance in the generation of \textit{SPARQL} queries from dialogues about knowledge graph facts and discuss insights about their individual capabilities as well as limitations. 
Our contributions include a benchmark study evaluating four \acp{llm}, utilizing both automatic metrics and human evaluation to identify eight common error types in generated graph queries, and a detailed discussion of prompting and fine-tuning strategies aimed at improving model performance. To ensure full reproducibility of our experiments, we have established a GitHub repository encompassing all model scripts, datasets, and evaluation outputs.\footnote{\href{https://github.com/sebischair/LLM-SP-CQA/}{https://github.com/sebischair/LLM-SP-CQA}}

\section{\uppercase{Related Work}}
Early semantic parsing methods were characterized by symbolic approaches rooted in production grammars and handcrafted linguistic features. With significant advancements driven by deep learning, there has been a shift towards neural approaches that cast semantic parsing as a machine translation problem by training neural networks that convert natural language input into a formal target language. Sequence-to-sequence neural networks established themselves as a general modeling framework consisting of an encoder and a decoder \cite{sutskever2014seq2seq}. The former encodes natural language utterances into hidden representations, whereas the latter decodes representations of the target formalism sequentially. 
The adaptability of neural networks eliminates the necessity of defining lexicons or manually crafted features, enabling models to generalize across various domains and meaning representation languages \cite{wang-etal-2020-rat}. Furthermore, scholars have developed hybrid semantic parsing approaches that combine symbolic and neural components to leverage advantages from both the good context representation obtained by neural nets and reduced decoding complexity and adherence to predefined structures due to constraints introduced by grammars \cite{lin-etal-2022-towards}.

While the majority of existing work focuses on parsing independent natural language utterances without considering broader contextual information, a growing body of literature is about contextualized semantic parsing that takes surrounding information beyond the current utterance into account, such as interaction histories \cite{li-etal-2020-context}. Therefore, context-aware parsing is particularly relevant for \ac{cqa} scenarios characterized by a series of interrelated utterances, ambiguous queries, and evolving search intents.

\begin{table*}
\footnotesize
\vspace{-0.2cm}
\caption{Overview of the two applied prompts. Parts marked with ``$< >$'' denote variables that are inserted at runtime based on the current test example. The complete few-shot prompt with three examples is provided in the linked repository.}\centering
\label{tab:prompts}
\renewcommand{\arraystretch}{0.75}
\begin{tabular}{|p{1.1cm}|p{13.85cm}|}
\hline
Prompt & Content\\
\hline
Zero-shot & \tiny \textcolor{violet}{SYSTEM:} Generate a SPARQL query that answers the given ’Input question:’. Use ’Entities:’, ’Relations:’ and ’Types:’ specified in the prompt to generate the query. The SPARQL query should be compatible with the Wikidata knowledge graph. Prefixes like ’wdt’ and ’wd’ have already been defined. No language tag is required. Use ’?x’ as variable name in the SPARQL query. Remember to provide only a SPARQL query in the response without any notes, comments, or explanations.
\\
& \tiny \textcolor{teal}{USER:} $<$conversation\_history$>$
\\
& \tiny Input question: $<$utterance$>$
\\
& \tiny Entities: $<$entities$>$
\\
& \tiny Relations: $<$relations$>$
\\
& \tiny Types: $<$types$>$
\\
\hline
Few- & \tiny [...]
\\
shot & \tiny \textcolor{teal}{USER:} Conversation history: 
\\
example & \tiny USER: Which administrative territory is the native country of Cirilo Villaverde? \\
& \tiny SYSTEM: \{'Q241': 'Cuba'\} \\
& \tiny Input question: Which is the national anthem of that administrative territory ? \\
& \tiny Entities: \{'Q241': 'Cuba'\} \\
& \tiny Relations: \{'P85': 'anthem'\} \\
& \tiny Types: \{'Q484692': 'hymn'\}
\\
& \tiny \textcolor{blue}{ASSISTANT:} SPARQL query: SELECT ?x WHERE \{ wd:Q241 wdt:P85 ?x . ?x wdt:P31 wd:Q484692 .  \}
\\
& \tiny [...]
\\
\hline
\end{tabular}
\end{table*}

Recent neural approaches have extended the sequence-to-sequence architecture to include contextual information, achieved by modifying either the encoder or decoder. Context-aware encoders adopt strategies like concatenating the current utterances with preceding ones \cite{zhang-etal-2019-editing} or only the most relevant utterances from the history \cite{liu2021how}. Decoders can incorporate context representations as supplementary input, often incorporating segments from previous queries \cite{suhr-etal-2018-learning}. Other scholars have also proposed hybrid methods that harness neural networks for contextualized representation learning while using grammar-based decoding \cite{guo-etal-2019-towards,liu2021how}.
Evaluating \ac{cqa} methods poses a considerable challenge, primarily due to the scarcity of available datasets constructed for this task. 
Since the few benchmarks for \ac{cqa} are often limited in scale, domain specificity, or dialogue length, our study takes advantage of the recently published dataset SPICE \cite{perez-beltrachini-etal-2023-semantic}, preventing benchmark leakage, where data from evaluation sets is occasionally used for \ac{llm} pre-training. Derived from the CSQA dataset \cite{saha2018complex}, an established benchmark for retrieval-based \ac{cqa}, SPICE extends CSQA by pairing dialogues with executable SPARQL queries along with answer triples from the Wikidata knowledge graph. Further details about this dataset are provided in Section~\ref{sec:experimental-setup}. 

The creators of the SPICE benchmark tested two strong baseline models and analyzed their performance across various question types. The first approach \textit{BertSP} adopts a standard sequence-to-sequence architecture for generating complete SPARQL queries \cite{gu2021beyond}. To tackle the extensive vocabulary of the Wikidata knowledge graph, dynamic vocabularies for entities and relations are derived from knowledge subgraphs corresponding to each question and its contextual information. This approach combines a BERT-based encoder \cite{devlin-etal-2019-bert}, fine-tuned specifically for semantic parsing, with a randomly initialized transformer network as the decoder. The second approach \textit{LasagneSP} adapts the \textit{LASAGNE} architecture, as proposed by \citet{kacupaj-etal-2021-conversational}. It employs an encoder-decoder transformer network to generate base logical forms (i.e., SPARQL templates), while a graph attention model is used to produce node representations by exploiting correlations between entity types and relations. An entity recognition module based on an inverted index is also part of the architecture. 

To the best of our knowledge, we are the first to evaluate the emergent capabilities of \acp{llm} that have not been explicitly trained for conversational semantic parsing. Unlike previously mentioned models, which chain multiple fine-tuned neural networks in a sequence, we apply in-context learning and post-processing. This enables the end-to-end generation of structured queries. We seek to investigate how \acp{llm} perform in understanding dialogues, resolving vocabularies, and generating SPARQL queries with correct syntax. Consequently, our objective transcends the scope of individual models; it strives for a comprehensive understanding of using \acp{llm} in conversational semantic parsing over knowledge graphs.

\section{\uppercase{Experimental Setup}}
\label{sec:experimental-setup}

\paragraph{Benchmark Dataset}
Our study aims to assess the performance of \acp{llm} in knowledge-based conversational search, with a particular focus on semantic parsing for \ac{cqa}. For the evaluation, we have chosen to use the \textit{SPICE} dataset from \citet{perez-beltrachini-etal-2023-semantic}. The dataset comprises conversational interactions between a user and an assistant. Each independent conversation is paired with SPARQL queries, which are executable against a knowledge graph engine to retrieve answers from a Wikidata snapshot. Furthermore, the dialogue transcripts within the SPICE dataset showcase conversational phenomena such as coreference, ellipsis, and clarifications. In some instances, clarifying questions and user responses accompany the SPARQL parse and query results. Obtaining correct SPARQL queries and corresponding answers requires handling a variety of different questions. The higher-order question types can be distinguished into logical reasoning, quantitative reasoning, comparative reasoning, verification, and simple questions. 

In total, the SPICE dataset consists of 197,000 dialogues, with an average of 9.5 conversation turns. Because the dataset only contains Wikidata references for entities, types, and relations, we carried out preprocessing steps to map references to their lexical forms. This was done to avoid relying on the models' intrinsic knowledge for resolving these lexical forms. For example, if an entity reference corresponds to the Wikidata ID Q30, we include the label ``United States of America'' through a lookup via the Wikidata.org website. From the more than 150,000 training examples, we constructed a smaller fine-tuning dataset with 30,000 conversations. These conversations contained between one and four independent conversational turns to simulate zero- and few-shot prompting, maintaining the same system message and prompt structure as during inference. 
Another preparation step was to sample down the test examples to a subset with 1,500 conversations. The reason for this was the resource constraint of keeping each model's required inference run time below 24 hours. To construct this test subset, we computed the distribution of the entire test set across all question categories and then determined the required samples for each category.

\paragraph{Models}
We compare four large language models of varying sizes with different prompting techniques. As a popular state-of-the-art \ac{llm} that is closed-source, we opted to include \textit{GPT-3.5-Turbo} (\textit{ChatGPT}) \cite{openai2022chat} in our comparison. It is optimized for dialogue interaction and has demonstrated remarkable zero-shot performance on various \ac{nlp} tasks and is often used as a benchmark for comparing \acp{llm}. We conducted our semantic parsing experiments with the model version GPT-3.5-Turbo-0613. Further, we decided to test \textit{LLaMA}, a collection of \acp{llm} developed and open-sourced by Meta \cite{touvron2023llama}. We include three model variations with 7B parameters of the first LLaMA version. In addition to the non-conversational base model, we included a fine-tuned model referred to as LoRA. The training was done through \textit{\ac{lora}} with 30,000 examples, a method that fine-tunes only a subset of the model's parameters, referred to as low-rank matrices, rather than updating the entire parameter space, improving the fine-tuning efficiency \cite{hu2022lora}. Another fine-tuned LLaMA model we tested is named \textit{Vicuna}, which was trained on a corpus of roughly 70,000 user-shared ChatGPT conversations crawled from the ShareGPT website \cite{chiang2023vicuna}. 

\begin{table}[b!]
\footnotesize
\caption{Semantic parsing performance for simple and complex questions evaluated by F1 score (F1), accuracy (ACC), and ratio of exact matches (EM). Bold values indicate the best performance for each metric.}
\renewcommand{\arraystretch}{0.75}
\label{tab:results}
\centering
\begin{tabular}{|l|c|c|c|c|}
  \hline
  & \multicolumn{2}{c|}{Zero-Shot} & \multicolumn{2}{c|}{Few-Shot} \\
  \cline{2-5}
  Model & F1 & EM & F1 & EM \\
  \hline
  \multicolumn{5}{|c|}{Simple Question (Direct)} \\
  \hline
  LLaMA-7B & 0.000 & 0.000 & 0.352 & 0.724 \\
  Vicuna-7B & 0.003 & 0.000 & 0.127 & 0.230 \\
  GPT-3.5-Turbo & 0.324 & 0.337 & 0.804 & 0.741 \\
  LoRA-7B & 0.867 & \textbf{0.970} & \textbf{0.963} & 0.917 \\
  LoRA-7B-512 & 0.867 & \textbf{0.970} & - & - \\
\hline
  \multicolumn{5}{|c|}{Simple Question (Coreference)} \\
  \hline
  LLaMA-7B & 0.000 & 0.000 & 0.350 & 0.568 \\
  Vicuna-7B & 0.000 & 0.000 & 0.189 & 0.321 \\
  GPT-3.5-Turbo & 0.491 & 0.234 & 0.636 & 0.623 \\
  LoRA-7B & 0.882 & 0.867 & 0.844 & 0.786 \\
  LoRA-7B-512 & \textbf{0.892} & \textbf{0.873} & - & - \\
\hline
  \multicolumn{5}{|c|}{Simple Question (Ellipsis)} \\
  \hline
  LLaMA-7B & 0.000 & 0.000 & 0.000 & 0.000 \\
  Vicuna-7B & 0.000 & 0.000 & 0.000 & 0.000 \\
  GPT-3.5-Turbo & 0.342 & 0.158 & 0.609 & 0.351 \\
  LoRA-7B & \textbf{0.855} & \textbf{0.754} & 0.618 & 0.526 \\
  LoRA-7B-512 & \textbf{0.855} & \textbf{0.754} & - & - \\
  \hline
  \multicolumn{5}{|c|}{Logical Reasoning (All)} \\
  \hline
  LLaMA-7B & 0.000 & 0.000 & 0.109 & 0.000 \\
  Vicuna-7B & 0.000 & 0.000 & 0.001 & 0.000 \\
  GPT-3.5-Turbo & 0.631 & 0.000 & \textbf{0.912} & 0.246 \\
  LoRA-7B & 0.900 & \textbf{0.926} & 0.810 & 0.779 \\
  LoRA-7B-512 & 0.900 & \textbf{0.926} & - & - \\
\hline
  \multicolumn{5}{|c|}{Comparative Reasoning (All)} \\
  \hline
  LLaMA-7B & 0.000 & 0.000 & 0.001 & 0.000 \\
  Vicuna-7B & 0.000 & 0.000 & 0.072 & 0.000 \\
  GPT-3.5-Turbo & 0.015 & 0.000 & 0.006 & 0.000 \\
  LoRA-7B & 0.000 & 0.000 & 0.001 & 0.000 \\
  LoRA-7B-512 & \textbf{0.315} & \textbf{0.114} & - & - \\
\hline
  & \multicolumn{2}{c|}{Zero-Shot} & \multicolumn{2}{c|}{Few-Shot} \\
  \cline{2-5}
   Model & ACC & EM & ACC & EM \\
\hline
  \multicolumn{5}{|c|}{Verification (Boolean) (All)} \\
  \hline
  LLaMA-7B & 0.000 & 0.000 & 0.000 & 0.000 \\
  Vicuna-7B & 0.000 & 0.000 & 0.365 & 0.162 \\
  GPT-3.5-Turbo & 0.000 & 0.000 & 0.926 & 0.480 \\
  LoRA-7B & \textbf{0.939} & 0.851 & 0.926 & 0.777 \\
  LoRA-7B-512 & \textbf{0.939} & \textbf{0.858} & - & - \\
  \hline
  \multicolumn{5}{|c|}{Quantitative Reasoning (Count) (All)} \\
  \hline
  LLaMA-7B & 0.000 & 0.000 & 0.152 & 0.027 \\
  Vicuna-7B & 0.000 & 0.000 & 0.091 & 0.008 \\
  GPT-3.5-Turbo & 0.197 & 0.008 & 0.485 & 0.212 \\
  LoRA-7B & \textbf{0.591} & \textbf{0.561} & 0.492 & 0.417 \\
  LoRA-7B-512 & \textbf{0.591} & \textbf{0.561} & - & - \\
  \hline
\end{tabular}
\end{table}

We set the token limit to 128 and the temperature parameter to 0, maximizing deterministic generation by favoring tokens with the highest probability. All models are prompted in the chat completion structure of the FastChat\footnote{\href{https://github.com/lm-sys/FastChat}{FastChat: https://github.com/lm-sys/FastChat}} platform, with a structured list of system, user, and assistant messages. Table~\ref{tab:prompts} displays the structure of each prompt. The main instruction is given as a system message. The user message contains the question, lexical forms of entities as well as relations, and the conversation history, which is created by including up to three last dialogue turns. The zero-shot prompt contains only a system message with a semantic parsing instruction. The few-shot prompt expands the instruction with three in-context examples of the task with different SPARQL constructs, such as ASK, SELECT, or COUNT. Furthermore, one example demonstrates using the conversation history to resolve an entity referenced from a previous conversational turn.

\section{\uppercase{Results and Discussion}}

\paragraph{Automatic Evaluation Results}

The performance metrics of the semantic parsing experiments for simple questions and complex questions are presented in Table~\ref{tab:results}. To ensure consistency in the evaluation, all metrics were computed on post-processed model predictions, which involved normalizing whitespace characters and removing ``SPARQL query:'' from the beginning of the generated output. Based on 1,500 conversations of the SPICE test dataset, we computed the exact match (EM) ratio, comparing the predicted queries to the ground truth queries. While F1 scores were calculated for question categories yielding sets of entities, the accuracy (ACC) metric was employed to measure performance in cases where the results constituted count or boolean values. The range for each metric lies between 0 to 1, with the optimal score being 1. The provided tables report results for 7 out of the 10 question types, focusing on those for which at least one model achieved a reasonable performance score above zero. Aside from the four \acp{llm}, we added results from a ``LoRA-7B-512'' model, which used a maximum token length of 512 instead of 128 tokens. This was done to assess if increasing the token limit could enhance the top-performing model.

\begin{table*}[h!]
\footnotesize
\vspace{-0.2cm}
\caption{Overview of the eight identified error types with examples from model generated predictions (PRED) and ground truth (GT) queries. Errors in the predictions are highlighted in red color.}
\renewcommand{\arraystretch}{0.75}
\label{tab:error-types} \centering
\begin{tabular}{|p{1.4cm}|p{2.6cm}|p{10.5cm}|}
  \hline
  Error Type & Definition & Example \\
  \hline
  Cutoff & \tiny PRED matches GT exactly but ends abruptly. & \tiny GT: [...] WITH \{ SELECT DISTINCT ?x (0 AS ?tupcount) WHERE \{ \{ \{ ?x wdt:P122 ?b . ?x wdt:P31 wd:Q7275 .  \} \} FILTER NOT EXISTS [...]  \newline
PRED: [...] WITH \{ SELECT DISTINCT ?x (0 AS ?tupcount) WHERE \{ \{ \{ ?x wdt:P122 ?b . ?x wdt:P31 \textcolor{red}{w}\\
\hline
  Deviating entities & \tiny PRED uses entity reference not specified in the prompt. & \tiny GT: SELECT DISTINCT ?x WHERE \{ ?x wdt:P101 ?y . VALUES ?y \{ wd:Q1622272 wd:Q170790 \}. ?x wdt:P31 wd:Q502895 .  \}  \}
 \newline PRED: SELECT ?x WHERE \{ ?x wdt:P101 wd:Q1622272 . ?x wdt:P101 wd:Q170790 . ?x wdt:P31 wd:\textcolor{red}{Q5} . \} \\
\hline 
  Alternative query & \tiny Alternative SPARQL query but correct result. & \tiny GT: SELECT ?x WHERE \{ wd:Q6177791 wdt:P451 ?x . ?x wdt:P31 wd:Q502895 .  \}
 \newline PRED: SELECT ?x WHERE \{ \textcolor{red}{?x wdt:P451 wd:Q6177791} . ?x wdt:P31 wd:Q502895 . \} \\
\hline  
  Incorrect result & \tiny Valid SPARQL query but incorrect result. & \tiny GT: SELECT ?x WHERE \{ wd:Q6177791 wdt:P451 ?x . ?x wdt:P31 wd:Q502895 .  \}   \newline
PRED: SELECT ?x WHERE \{ ?x wdt:P451 ?p . ?p wdt:Q502895 ?type . ?type wdt:commonName ?x . \} \\
\hline
  Language filter & \tiny PRED contains language filter. & \tiny GT: SELECT ?x WHERE \{ wd:Q123179 wdt:P69 ?x . ?x wdt:P31 wd:Q163740 .  \}
 \newline PRED: SELECT ?x WHERE \{ ?x wdt:P69 ?y . \textcolor{red}{FILTER (LANG( ?y)='en')} . \} LIMIT 1 \\
\hline
  Namespace definition & \tiny PRED uses name-spaces instead of wd and wdt & \tiny GT: SELECT ?x WHERE \{ [...]  \newline
PRED: \textcolor{red}{PREFIX wdt: \textless http://www . wikidata . org/entity/\textgreater PREFIX wd:\textless http://www . wikidata . org/prop/direct/\textgreater} SELECT ?x WHERE \{ [...] \\
\hline
  Off-prompt & \tiny PRED is unrelated to prompt and contradicts desired output format.  & \tiny GT: SELECT ?x WHERE \{ wd:Q23487488 wdt:P702 ?x . ?x wdt:P31 wd:Q863908 .  \}  \newline
PRED: \textcolor{red}{Input question: What is the nucleic acid sequence that is encoded by 16S rRNA methyltransferase GidB SSA\_0605 ? Entities: \{'Q23487488': '16S rRNA methyltransferase} [...] \\
\hline
  Syntax \break error & \tiny PRED is invalid SPARQL & \tiny GT: SELECT DISTINCT ?x WHERE \{ ?x wdt:P166 ?y . VALUES ?y \{ wd:Q918055 wd:Q133160 wd:Q920783 \}. ?x wdt:P31 wd:Q502895 .  \}  \newline
PRED: SELECT ?x WHERE \{ \textcolor{red}{?x wdt:P166 ?award ?award} wdt:Q918055 ?award wdt:Q133160 ?award wdt:Q920783 \} \\
\hline
\end{tabular}
\end{table*}

Upon inspecting the metrics for simple questions in Table~\ref{tab:results}, it becomes evident that LLaMA and Vicuna show the worst performance, particularly with zero-shot prompting, where they fail to produce valid queries regardless of the question type. Although the provision of in-context examples significantly improves their performance on direct and coreference questions, achieving F1 scores of up to 0.352 for LLaMA and 0.189 for Vicuna, few-shot prompting does not extend to their ability to handle questions that involve ellipsis. The GPT-3.5-Turbo model demonstrates superior performance compared to LLaMA and Vicuna, being capable of parsing queries in zero-shot settings and effectively addressing questions with an ellipsis. Notably, the fine-tuned LoRA model consistently surpasses all models, producing exact ground truth SPARQL queries with an ACC ranging from 75\% to 97\%. It can be observed that LoRA often performs better without few-shot prompting. We hypothesize that the examples of the few-shot prompt might introduce a superfluous information bias since the model already learned from task-specific examples during fine-tuning. When dealing with dialogue phenomena such as coreferences (i.e., linguistic expressions like pronouns referring back to entities mentioned in a previous turn) and ellipsis (i.e., omission of one or more words for brevity because they can be inferred from the dialogue context), generating parses that precisely match the ground truth proves challenging for all \acp{llm}. This aligns with observations from \citet{perez-beltrachini-etal-2023-semantic}, where the SPICE baseline models also struggled with these two phenomena. Still, the fine-tuned LoRA model handles these complexities well, achieving similar F1 scores across all simple question types.

Concerning the \acp{llm}' performance metrics on more complex question types, semantic parsing proves to be more difficult. Complex questions require a number of logical and numerical operations over entity sets associated with longer SPARQL parses (e.g., \textit{How many bodies of water or watercourses are situated nearby Lübeck?}). LLaMA and Vicuna exhibit inferior performance compared to simple questions, with the exception of verification questions that result in a boolean value. The latter is the only category where Vicuna outperforms LLaMA with an F1 score of 0.365. The substantially larger GPT-3.5-Turbo model excels in logical and verification questions in few-shot scenarios, even though it took few-shot examples to get the model to use the ASK instead of the SELECT operator for verification questions that should return a boolean value. The exceptional performance could be attributed to the model's explicit training on tasks that involve logical reasoning operations. Another interesting observation about these question types is that GPT-3.5-Turbo demonstrates the ability to infer parses that yield correct results, achieving comparable F1 scores to LoRA, even though the EM ratio is considerably lower, suggesting that it has learned to convey the same question intent through an alternative SPARQL expression. 

Overall, LoRA emerges again as the best-performing \ac{llm}, producing the highest number of EM queries; however, for the most complex question types, such as quantitative and comparative reasoning, its performance is limited, akin to the other models. It is worth noting that using LoRA with 512 instead of 128 maximum tokens leads only to better performance on comparative reasoning questions. This indicates that the model successfully generated correct outputs for a few queries but often terminated abruptly upon reaching the token limit, which, in turn, resulted in syntax errors. Given that the ground truth queries for comparative reasoning are relatively lengthy, extending the maximum token limit even further could yield enhancements in performance.

\paragraph{Human Evaluation Results}


Besides our metric-based performance assessment of semantic parsing, we carried out a qualitative human evaluation to get further insights into the \acp{llm}' output. Two researchers manually analyzed a sample of 15 generated queries for each of the 10 question types, resulting in the examination of a total of 150 predictions, although six ground truth queries were absent from the SPICE dataset and were thus excluded from the analysis. By employing an iterative process involving the creation and consolidation of error categories, we successfully identified eight prevalent error types, as delineated in Table~\ref{tab:error-types}. For each error type, we provide a short definition accompanied by an example to juxtapose the ground truth query with the erroneous generated output. For instance, the \acp{llm} sometimes ignored the instruction given in the prompt. In other cases, they included entities that were wrong or not specified previously, cut off abruptly in the middle of the query, or generated parses with syntactical errors. 

To gain a more profound understanding of the error occurrence rates specific to each model and prompt combination, we present the relative frequencies of these error types in Table~\ref{tab:error-frequencies}. 
Many of these errors manifested in the predictions generated by LLaMA and Vicuna. Outputs from GPT-3.5-Turbo and LoRA exhibited a higher degree of reliability and a diminished incidence of such errors. Vicuna, GPT-3.5-Turbo, and LoRA demonstrate the ability to generate zero-shot output that aligns with the desired output written in the prompt. This outcome is consistent with expectations for instruction-tuned and fine-tuned models, suggesting their efficacy in aligning with user instructions and prompts. However, all outputs produced by the LLaMA zero-shot model are off-prompt (1.00), meaning that they did not contain a SPARQL query as the only desired output format, suggesting that a textual description of a complex task without including in-context examples is insufficient for LLaMA. An intriguing finding is that the issue of off-prompt errors can be effectively mitigated in all models by including SPARQL examples within the prompt, thus enhancing model performance and alignment with user intent. It is worth mentioning that across all \acp{llm}, 0.10 is the lower bound for the corresponding relative frequencies. The reason for this observation is the clarification question type. Because our study exclusively focuses on semantic parsing, we do not consider clarifying questions in the instructions leading to off-prompt behavior.

Output classified as an \textit{incorrect result} represents the relative frequency of a syntactically valid generated query that retrieved the wrong result (e.g., boolean, entity set, or integer). Within the few-shot setting, Vicuna demonstrated the least favorable performance, with an error rate of 0.86, followed by LLaMA (0.82) and GPT-3.5-Turbo (0.63). In contrast, LoRA produced queries with the fewest incorrect results, with a ratio of 0.20. Notably, in the case of LoRA, introducing few-shot examples resulted in nearly double the number of incorrect results compared to its zero-shot performance (0.12). This phenomenon suggests that the inclusion of few-shot examples may exert a negative bias on the already fine-tuned LoRA model.

\begin{table*}[h]
\footnotesize
\vspace{-0.2cm}
\caption{Relative frequency of error categories for zero-shot and few-shot prompts in the evaluated sample of 150 predictions. The asterisk symbols denote: ``*'' excluding off-prompt predictions, ``**'' excluding off-prompt and cutoff predictions, and ``***'' excluding off-prompt and syntax error predictions.}
\renewcommand{\arraystretch}{0.75}
\label{tab:error-frequencies} 
\centering
\begin{tabular}{|l|c|c|c|c|}
  \hline
  Error Type & LLaMA-7B & Vicuna-7B & GPT-3.5-Turbo & LoRA-7B \\
  \hline
   \multicolumn{5}{|c|}{Relative error frequency: zero-shot / few-shot} \\
  \hline
  Cutoff & - / - & - / - & - / - & 0.33 */ 0.24 * \\
  Deviating entities & - / 0.05 * & 0.04 */ 0.02 * & 0.06 */ 0.03 * & 0.02 */ 0.04 * \\
  Alternative query & - / - & - / - & 0.08 / 0.06 & 0.01 / 0.01 \\
  Incorrect result & - / 0.82 *** & 0.97 *** / 0.86 *** & 0.69 *** / 0.63 *** & 0.12 *** / 0.20 *** \\
  Language filter & - / - & 0.33 */ 0.06 * & 0.07 */ 0.02 & - / - \\
  Namespace definition & - / - & 0.11 */ - & - / - & - / - \\
  Off-prompt & 1.00 / 0.10 & 0.13 / 0.10 & 0.10 / 0.10 & 0.10 / 0.10 \\
  Syntax error & - / 0.16 ** & 0.71 **/ 0.26 ** & 0.20 **/ 0.17 ** & 0.01 **/ 0.10 ** \\
  \hline
\end{tabular}
\end{table*}

A similar pattern of few-shot behavior becomes evident when assessing queries with \textit{syntax errors}. This type was used to classify non-executable queries. Except for the LoRA model, few-shot prompting improves the ratio of syntactically valid queries compared to zero-shot prompting. GPT-3.5-Turbo (0.17) is significantly better than Vicuna (0.26) in both zero-shot and few-shot scenarios, while LLaMA (0.16) few-shot achieves a very similar occurrence rate as GPT-3.5-Turbo. LoRA generates the smallest number of syntax mistakes with few-shot (0.10) and even less in zero-shot prompting (0.01). We hypothesize that this is due to its exposure to 30,000 examples of correct SPARQL queries during fine-tuning. 

Another common error type pertains to \textit{deviating entities}, wherein Wikidata references are used in the predicted query without being explicitly specified as part of the prompt. This error type has a uniform relative frequency across all model-prompt combinations. Looking at these errors more closely, we see cases in which parts of the original reference are omitted, for example, using the Wikidata ID Q5 instead of the provided one Q502895. Moreover, two analyzed samples offer limited information regarding the conversation history of the prompt, leading to models hallucinating other entities or relation references when the information is unavailable. This issue could potentially be alleviated by including references for all relevant entities, types, and relations within the full conversation history of the prompt. In the system prompt, the \acp{llm} are explicitly instructed to only use specified entities, relations, and types. 

Furthermore, we instructed the models to refrain from defining namespace prefixes and to use the Wikidata internal prefixes ``wdt'' and ``wd'' instead. Considering errors with \textit{namespace definitions}, we measure the relative frequency where the model does not follow that instruction. Only Vicuna (zero-shot) shows this undesirable behavior with a frequency of 0.11. Similar to this error type, we also analyze how well the model follows the system instruction to refrain from using \textit{language filters}. The SPICE knowledge graph only contains English triples, whereas no language labels are provided. Consequently, filtering in the query for a language does not yield any results, even if it is syntactically valid. Therefore, in the system prompt, we specify that the generated query should not filter for languages. This error type was only observed in generations from Vicuna and GPT-3.5-Turbo, with few-shot prompting leading to significant improvements over zero-shot. In-context examples reduced the error for Vicuna from 0.33 to 0.06 and GPT-3.5-Turbo from 0.07 to 0.02.

A prediction is considered to be \textit{cutoff} if it matches the expected output, but the generation stops abruptly before completing the query. The maximum token length, a hyperparameter for \acp{llm}, is the cause of this issue. Increasing it can mitigate the problem. Regarding the analyzed model-prompt combinations, such errors are only found in predictions from LoRA. All models except LoRA deviate from the ground truth query before reaching maximum tokens. To see if we can improve the generations for LoRA, we experimented with increasing the limit from 128, which we use as standard for all model-prompt configurations, to 512. Although the four times higher limit improves the performance as shown in Table~\ref{tab:results}, the token limit is only reached for very complex question types, such as comparative reasoning questions.

Lastly, the error type named \textit{alternative query} was used to classify predictions that constitute valid SPARQL queries that yield a correct result, albeit not exactly matching with the ground truth query from the SPICE dataset. Hence, this error type reduces the EM performance while leaving the ACC and F1 scores unaffected. As presented in Table~\ref{tab:error-frequencies}, the generation of alternative queries was only observed in outputs from GPT-3.5-Turbo and LoRA. The GPT-3.5-Turbo model, specifically in the zero-shot setting, produced the highest number of instances within this category (0.08), with few-shot prompting decreasing it further (0.06). We assume that this may be attributed to GPT-3.5-Turbo's extensive pre-training, which likely equipped it with a deeper understanding of SPARQL queries, enabling it to formulate alternative queries that still yield the correct results. Conversely, LoRA generated this type of substitute query in merely 1\% of the analyzed outputs, with no discernible difference between zero-shot and few-shot prompting settings.

\paragraph{Discussion}

Through our study's experimental results, we gained several valuable insights into how \acp{llm} perform in semantic parsing for \ac{cqa}. Each model we evaluated demonstrated at least a degree of proficiency in generating structured SPARQL queries, even if they were not explicitly trained for this specific task. Some \acp{llm} showed the ability to handle coreference and ellipsis within the context of simple questions. This aptitude indicates their capacity to leverage contextual information from the dialogue histories to produce correct SPARQL queries from ambiguous user questions. Nevertheless, when faced with these linguistic phenomena, especially in more complex question types, the \acp{llm}' performance experienced a significant decrease.


When analyzing overall performance as a weighted average score comprised of ACC and F1 scores across all question types, LLaMA base model demonstrates almost twice as good performance as the fine-tuned Vicuna model. This may suggest that fine-tuning on conversational data, as in the case of Vicuna, might have a trade-off, potentially leading to a decrease in generative capabilities concerning structured query languages. It is noteworthy that LLaMA and Vicuna failed to generate valid SPARQL queries in the zero-shot scenario. 
The significantly larger GPT-3.5-Turbo model outperformed LLaMA and Vicuna with zero- and few-shot prompting. We found GPT-3.5-Turbo's ability to generate alternative queries especially interesting. Although these queries did not match the ground truth query, they still managed to return correct results, which may be attributed to extensive pre-training on documents containing structured, formal languages, equipping the model with substantial knowledge of SPARQL. Our fine-tuned 7B parameter LoRA model surpassed the performance of the considerably larger GPT-3.5-Turbo  model. Our analysis of common errors also revealed that LoRA consistently generated the fewest errors across the identified error types. The top-performing model, LoRA-7B-512, attains an overall weighted average performance of 0.724, falling short of the best baseline model in the SPICE paper, BertSP$_G$, which scores 0.815 \cite{perez-beltrachini-etal-2023-semantic}, although it is worth noting that BertSP$_G$ was trained with five times more examples. Also, it should be reiterated that we used a subset of the test data, as detailed in Section~\ref{sec:experimental-setup}, so direct comparisons have to be made with caution.

The experimental results highlight the effectiveness of few-shot prompting in reducing errors and increasing performance metrics in all models except for LoRA. Errors related to off-prompt and wrongly formatted output saw the most significant improvements. LoRA was working best in the zero-shot setting, as pointed out before. We assume that a model like LoRA, which has previously been fine-tuned with in-context examples, might not benefit from them further; in fact, it could be negatively impacted by biasing the generation process. Apart from few-shot prompting, employing rule-based approaches could further minimize prediction errors. These strategies might involve syntax checking, utilizing entity dictionaries, checking for unwanted language filters, and removing natural language output that is not SPARQL.

Finally, it is important to acknowledge that our study has certain limitations. We have concentrated on semantic parsing of queries from dialogues, although we recognize the importance of exploring other tasks, such as extracting triples or constructing subgraphs in different graph languages. We also suggest further extending our foundational evaluation by additional human assessments and including a wider array of recently published models, especially those trained on program code or structured data documents. Moreover, the SPICE dataset is limited to English. Since pre-training corpora of \acp{llm} primarily consist of English text data, they likely work better where entities and relations correspond to meaningful English words. Consequently, it is to be expected that \acp{llm} exhibit worse performance on benchmarks with more morphologically rich languages.

\section{Conclusion}
We compared \acp{llm} in conversational semantic parsing. Our findings indicate that even smaller, fine-tuned 7B-\acp{llm} exhibit reasonable performance in generating SPARQL queries from dialogues, although they might not always be syntactically valid or yield the correct result. We also discussed model-specific differences and common errors that can be mitigated through few-shot prompting and fine-tuning. In future work, we intend to delve into the applicability of our findings to different query languages. Further, we plan to conduct user evaluations of deployed \ac{llm}-based \ac{cqa} systems for practical search scenarios.

\section*{\uppercase{Acknowledgements}}
This work has been supported by the German Federal Ministry of Education and Research grant 01IS17049.


\bibliographystyle{apalike}
{\small
\bibliography{example_acro}}

\end{document}